# Exploring the Statistical Derivation of Transformational Rule Sequences for Part-of-Speech Tagging

**Lance A. Ramshaw**
Inst. for Research in Cognitive Science
University of Pennsylvania
3401 Walnut Street #412–C
Philadelphia, PA 19104–6228 USA
ramshaw@linc.cis.upenn.edu
and
Dept. of Computer Science
Bowdoin College
Brunswick, ME 04011 USA

**Mitchell P. Marcus**
Computer and Information Science Dept.
University of Pennsylvania
558 Moore Building
Philadelphia, PA 19104–6389 USA
mitch@linc.cis.upenn.edu

## Introduction

Eric Brill in his recent thesis (1993b) proposed an approach called "transformation-based error-driven learning" that can statistically derive linguistic models from corpora, and he has applied the approach in various domains including part-of-speech tagging (Brill, 1992; Brill, 1994) and building phrase structure trees (Brill, 1993a). The method learns a sequence of symbolic rules that characterize important contextual factors and use them to predict a most likely value. The search for such factors only requires counting various sets of events that actually occur in a training corpus, and the method is thus able to survey a larger space of possible contextual factors than could be practically captured by a statistical model that required explicit probability estimates for every possible combination of factors. Brill's results on part-of-speech tagging show that the method can outperform the HMM techniques widely used for that task, while also providing more compact and perspicuous models.

Decision trees are an established learning technique that is also based on surveying a wide space of possible factors and repeatedly selecting a most significant factor or combination of factors. After briefly describing Brill's approach and noting a fast implementation of it, this paper analyzes it in relation to decision trees. The contrast highlights the kinds of applications to which rule sequence learning is especially suited. We point out how it manages to largely avoid difficulties with overtraining, and show a way of recording the dependencies between rules in the learned sequence. The analysis throughout is based on part-of-speech tagging experiments using the tagged Brown Corpus (Francis and Kucera, 1979) and a tagged Septuagint Greek version of the first five books of the Bible (CATSS, 1991).

## Brill's Approach

This learning approach starts with a supervised training corpus and a baseline heuristic for assigning initial values. In the part-of-speech tagging application, for example, the baseline heuristic might be to assign each known but ambiguous word whatever tag is most often correct for it in the training corpus, and to assign all unknown words an initial tag as nouns. (Brill's results point out that performance on unknown words is a crucial factor for part-of-speech tagging systems. His system is organized in two separate rule sequence training passes, with an important purpose of the first pass being exactly to predict the part-of-speech of unknown words. However, because the focus in these experiments is on understanding the mechanism, rather than on comparative performance, the simple but unrealistic assumption of a closed vocabulary is made.)

The learner then works from those baseline tag assignments using a set of templates that define classes of transformational rules, where each rule changes some assigned values based on characteristics of the neighborhood. Again, for tagging, the rule templates typically involve either the actual words or the tags currently assigned to words within a few positions on each side of the value to be changed. The rule templates used in these experiments involve up to two of the currently-assigned tags on each side of the tag being changed; they include [ — C A/B — — ] (change tag A to tag B if the previous tag is C) and [ — — A/B C D ] (change A to B if the following two tags are C and D). During training, instantiated rules like [ — DET V/N — — ] are built by matching these templates against the training corpus.

A set of such templates combined with the given part-of-speech tagset (and vocabulary, if the rule patterns also refer directly to the words) defines a large space of possible rules, and the training process operates by using some ranking function to select at each step a rule judged likely to improve the current tag assignment. Brill suggests the simple ranking function of choosing (one of) the rule(s) that makes the largest net improvement in the current training set tag assignments. Note that applying a rule at a location can have a positive effect (changing the current tag assignment from incorrect to correct), a negative one (from correct to some incorrect value), or can be a neutral move (from one incorrect tag to another). Rules with the largest positive minus negative score cause the largest net benefit. In each training cycle, one such rule is selected and applied to the training corpus and then the scoring and selection process is repeated on the newly-transformed corpus. This process is continued either until no beneficial rule can be found, or until the degree of improvement becomes less than some threshold. The scoring process is tractable in spite of the huge space of possible rules because rules that never apply positively can be ignored.

The final model is thus an ordered sequence of pattern-action rules. It is used for prediction on a test corpus by beginning with the predictions of the baseline heuristic and then applying the transformational rules in order. In our test runs, seven templates were used, three templates testing the tags of the immediate, next, and both neighbors to the left, three similar templates looking to the right, and a seventh template that tests the tags of the immediate left and right neighbors. The first ten rules learned from a training run across a 50K-word sample of the Brown Corpus are listed in Fig. 1; they closely replicate Brill's results (1993b, page 96), allowing for the fact that his tests used more templates, including templates like "if one of the three previous tags is A".

Brill's results demonstrate that this approach can outperform the Hidden Markov Model approaches that are frequently used for part-of-speech tagging (Jelinek, 1985; Church, 1988; DeRose, 1988; Cutting et al., 1992; Weischedel et al., 1993), as well as showing promise for other applications. The resulting model, encoded as a list of rules, is also typically more compact and for some purposes more easily interpretable than a table of HMM probabilities.

## An Incremental Algorithm

It is worthwhile noting first that it is possible in some circumstances to significantly speed up the straightforward algorithm described above. An improvement in our experiments of almost two orders of magnitude (from four days to under an hour) was achieved by using an incremental approach that maintains lists of pointers to link rules with the sites in the training corpus where they apply, rather than scanning the corpus from scratch each time. The improvement is particularly noticeable in the later stages of training, when the rules being learned typically affect only one or two sites in the training corpus. Note, however, that the linked lists in this incremental approach require a significant amount of storage space. Depending on the number of possible rules generated by a particular combination of rule templates and training corpus, space constraints may not permit this optimization.

Incrementalizing the algorithm requires maintaining a list for each rule generated of those sites in the corpus where it applies, and a list for each site of the rules that apply there. Once one of the highest-scoring rules is selected, its list of site pointers is first used to make the appropriate changes in the current tag values in the corpus. After making the changes, that list is used again in order to update other rule pointers that may have been affected by them. It suffices to check each site within the span of the largest defined rule template from each changed site, testing to see whether all of its old rule links are still active, and whether any new rules now apply at that site. Our current algorithm is shown in Fig. 2. Note that, after the initial setup, it is necessary to rescan the corpus only when updating uncovers a rule that has not previously had any positive effect.

## Rule Sequences and Decision Trees

To understand the success of Brill's new method, it is useful to compare it with the decision tree approach (Breiman et al., 1984; Quinlan, 1993), which is an established method for inducing compact and interpretable models. The key difference is that decision trees are applied to a population of non-interacting problems that are solved independently, while rule sequence learning is applied to a sequence of interrelated problems that are solved in parallel, by applying rules to the entire corpus. The following sections discuss how this parallel approach allows leveraging of partial solutions between neighboring instances, but also requires that the rules themselves be largely independent. While decision trees can synthesize complex rules from simple tests, rule sequence learning requires those combinations to be built into the templates.

### Leveraged Learning

Decision trees are traditionally applied to independent problem instances encoded as vectors of measurements for the various possibly-relevant factors. In predicting the part of speech of a word in a corpus, such factors would include the identities of the neighboring words within some window. However, it would also be useful to know the currently predicted tags for those words, since the tag-assignment problems for neighboring words in a corpus are not independent. The rule sequence learning technique is particularly well adapted to a corpus that is inherently a sequence of such interrelated problem instances. Because the rule patterns in a part-of-speech system do depend in part on the unknown part-of-speech values at neighboring locations,

| Pass | | | Rule | | | Pos. | Neg. | Neut. |
|------|---|---|------|---|---|------|------|-------|
| 1.   | — | — | TO/IN  | AT     | — | 227 | 0  | 0 |
| 2.   | — | TO | NN/VB | —     | — | 113 | 13 | 0 |
| 3.   | — | — | TO/IN  | NN     | — | 49  | 0  | 0 |
| 4.   | — | IN | PPS/PPO | —    | — | 51  | 4  | 0 |
| 5.   | — | — | TO/IN  | NP     | — | 46  | 0  | 0 |
| 6.   | — | — | TO/IN  | PP$    | — | 46  | 1  | 0 |
| 7.   | — | — | CS/DT  | NN     | — | 52  | 11 | 1 |
| 8.   | — | HVD | VBD/VBN | —   | — | 38  | 0  | 0 |
| 9.   | — | — | CS/QL  | —      | CS | 41 | 7  | 0 |
| 10.  | — | MD | NN/VB  | —     | — | 32  | 0  | 0 |

Figure 1: First 10 Rules Learned on Brown Corpus Sample

```
// Records for locations in the corpus, called "sites",
// include a linked list of the rules that apply at that site.
// Records for rules include score components (positive, negative, and neutral)
// and a linked list of the sites at which the rule applies.
// A hash table stores all rules that apply positively anywhere in the training.

scan corpus using templates, making hash table entries for positive rules
scan corpus again to identify negative and neutral sites for those rules
loop
    high_rule := some rule with maximum score
    if high_rule.score <= 0
        then exit loop
    output rule trace
    for each change_site on high_rule.site_list do
        apply high_rule at change_site by changing current tag
    unseen_rules := ∅
    for each change_site on high_rule.site_list do
        for each test_site in the neighborhood of change_site do
            new_rules_list := NIL
            for each template do
                if template applies at test_site
                    then add resulting rule to new_rules_list
            for each rule in test_site.rules_list − new_rules_list do
                remove connection between rule and test_site
            for each rule in new_rules_list − test_site.rules_list do
                if rule in hash table
                    then make new connection between rule and test_site
                    else unseen_rules := unseen_rules ∪ {rule}
    if unseen_rules ≠ ∅ then
        add unseen_rules to hash table
        for each site in corpus do
            for each rule in unseen_rules do
                if rule applies at site then
                    make connection between rule and site
                    adjust appropriate rule score (positive, negative, or neutral)
end loop
```

Figure 2: Incremental Version of Rule Sequence Learning Algorithm

it seems useful to allow those patterns to be based at each point on the system's best current guess for those values. It is difficult to take account of that kind of dependence in a traditional decision tree, since changes in neighboring tag predictions can force the recomputation of predicate splits higher in the tree. Breaking the tag prediction process up into a series of rules that can each be applied immediately to the entire corpus is a simple scheme that allows for that kind of leverage. Much as when a bank compounds interest, this allows the system to base its future learning on the improved estimates of neighborhood tags resulting from the operation of earlier rules.

A non-leveraged learner would have to build rules or trees based only on the unchanging features of the neighboring words and perhaps the baseline guesses of their tags. In effect, such a learner would be forced to try to resolve the ambiguity at the neighboring location as part of the rule for the primary site, using as evidence only cases where the two occur together. The leveraging approach allows the system to factor the best current guess for the neighboring site in terms of all the evidence into the choice for the primary site. It is to allow for leveraging that the model is formulated as a sequence of individual rules.

### Largely Independent Rules

This breaking up of the rule sequence model into largely independent rules also results in another important difference between rule sequence learning and decision trees. In the building of a decision tree, an elementary predicate is selected at each step to split a single leaf node, meaning that it is applied only to those training instances associated with that particular branch of the tree. The two new leaves thus created effectively represent two new classification rules, each one selecting exactly the instances that classify to it, and thus each including all of the predicates inherited down that branch of the tree. In the rule sequence method, on the other hand, the rules are generated from the templates as they are applied to the whole corpus in a largely independent manner; there is no corresponding inheritance of earlier predicates down the branches of a tree.

Note that one could simulate the decision tree style in a sequence learner by adding to the pattern for each rule template a variable-length field that records the complete history of rules which have affected that location. Then, as in a decision tree, a rule generated at one site in the training set would be scored only against sites whose previous rule history exactly matched its own. But rule sequence learning as defined here is not sensitive in that way to the previous rule history.

The "largely independent" rules in the sequence would be fully independent if the system were not doing leveraging; if all rule patterns were tested each time against the original baseline tag predictions, then there would be no way for earlier rules to affect later ones in the sequence. Leveraging does make later rules dependent on the results of earlier ones, but it does so to a strictly limited degree, which is generally much weaker than the direct inheritance of rules down decision tree branches.

To see the limitation, suppose that templates could test the current tag of the word to be changed, but could only consult the baseline tags for the rest of the pattern. Earlier rule firings could then affect what rules might later apply at a particular location only by changing the current tag assignment for that location itself to one of the other possible tag values. Each rule firing would make potentially applicable at the locations affected all rules whose central pattern element specify that new tag value, while disabling those rules whose patterns specify the old value. The training set at any time during training would thus in effect be partitioned for purposes of rule application into at most as many classes as there are tags. Such a system can be pictured as a lattice with one column for each tag assignment and with a single slanting arc at each generation that moves some corpus locations from one column to another.

While a decision tree path can encode an arbitrary amount of information in its branching, this system is forced to merge as often as it branches, which requires the rules to be more independent. Furthermore, the system's ability to use even the available partitioning in order to construct dependent rule sequences is further limited, since tag changes are only made when some subset of the data is identified for which the new tag is more representative of the training corpus; the learner is not free to use tag assignments to encode arbitrary rule dependencies. Even in the actual system, where the leveraging can include changes in the neighborhood as well as at the location itself, the rule sequence mechanism still appears to have much less power to create complex combined rules than do decision trees.

Because rule sequence learners are more limited in terms of the connections between rules that they can construct during training, they must begin with more complex predicates built into their rule templates. If the templates in a rule sequence run are not strong enough to distinguish the important patterns in the data, performance will naturally suffer. But if the rule templates that are likely to be useful can be predicted in advance, the rule sequence approach can benefit both from leveraging and, as shown later, from decreased fragmentation of the training set.

### Scoring Metrics

This difference in organization between rule sequence learning and decision trees carries through naturally to the scoring methods used to select the next rule to apply. Decision trees often select the split which most reduces either a diversity index or some measure based on the conditional entropy of the truth given the tree's predictions (Breiman et al., 1984; Quinlan and Rivest, 1989; Quinlan, 1993). Note that these metrics may select a split that does not change the score of the current predictions against the truth, for instance by splitting a node in such a way that both children still have the same plurality class as the parent. Such a split may still make sense in entropy terms if the distributions of the other tags in the two new nodes are substantially different, thus suggesting that later rules will have an easier time isolating particular tag values. In a rule sequence learner, however, there is less likely to be any advantage to such a split, since the instances whose tags are changed by that rule will then be mixed with others that were already assigned the new tag for other reasons. The net benefit metric that is actually used in rule sequence learning is equivalent in decision tree terms to using the resubstitution estimate of the misclassification rate. While that metric is not ideal for decision trees, it appears to work well for rule sequence learning, where the mechanism is strictly limited in terms of the connections between rules that it can construct.

### Overtraining

It is particularly interesting to compare rule sequences with decision trees in terms of the risk of overtraining (or "overfitting"). One of the intriguing features of rule sequence learning is its apparent resistance to overtraining. For example, Fig. 3 shows the graph of percent correct on both training set (solid line) and test set (dotted line) as a function of the number of rules applied for a typical part-of-speech training run on 120K words of Greek text. The training set performance naturally improves monotonically, given the nature of the algorithm, but the surprising feature of that graph is that the test set performance also improves monotonically, except for minor noise, and this seems to be true for the great majority of our rule sequence training runs. This is in marked contrast to similar graphs for decision trees or neural net classifiers or for the iterative EM training of HMM taggers on unsupervised data, where performance on the test set initially improves, but later significantly degrades.

Experiments suggest that part of the difference is due to knowledge embodied in the templates. When a part-of-speech training run is supplied with relevant templates, as in Fig. 3, one gets an "improve to plateau" test-set curve. Irrelevant templates, however, can lead to overtraining. Fig. 4 shows that noticeable overtraining results from using just a single irrelevant template, in this case, one that tested the tags of the words five positions to the left and right, which seem likely to be largely uncorrelated with the tag at the central location.

Fig. 5, where the single irrelevant template is combined with the seven normal templates, shows that in such cases, most of the overtraining happens late in the training process, when most of the useful relevant templates have already been applied. At that stage, as always, the templates are applied to each remaining incorrectly-tagged site, generating candidate rules. Each rule naturally succeeds at the site that proposed it, but most are now effectively random changes, which are thus likely to do more harm than good when tried elsewhere, especially since most of the assigned tags at this stage are correct. Thus if the rule's pattern matches elsewhere in the training set, it is quite likely that the change there will be negative, so that the unhelpful rule will not be learned. Thus the presence of relevant templates supplies an important degree of protection against overtraining from any irrelevant templates, both by reducing the number of incorrect sites that are left late in training and by raising the percentage already correct, which makes it more likely that bad rules will be filtered out. The same applies, of course, to relevant and irrelevant instances of mixed templates, which is the usual case.

Most of the overtraining will thus come from patterns that match only once in the training set (to their generating instance). Under these assumptions, note that applying a score threshold $> 1$ can significantly reduce the overtraining risk, just as decision trees sometimes control that risk by applying a threshold to the entropy gain required before splitting a node. Brill's system uses a score threshold of 2 as the default, thus gaining additional protection against overtraining, while our experimental runs have been exhaustive, in order to better understand the mechanism.

Using test runs like those plotted above for irrelevant templates of various degrees of complexity, we also found a connection in terms of overtraining risk between the inherent matching probability of the templates used and the size of the training set. A large training set means a larger number of incorrect sites that might engender overtrained rules, but also a better chance of finding other instances of those rule patterns and thus filtering them out. The combination of those factors appears to cause the risk of overtraining for a particular irrelevant template to first rise and then fall with increasing training set size, as the initial effect of increased exposure is later overcome by that of increased filtering from further occurrences of the patterns.

In comparing this with decision trees, the key contrast is that the filtering effect there decreases as training proceeds. The splitting predicates there are applied to increasingly small fragments of the training set, so that the chance of filtering counterexamples also decreases. (To put it in decision tree terms, with few points left in the rectangle being split, it becomes more likely that an irrelevant predicate will incorrectly appear to provide a useful split.) But since rule sequence learning continues to score its essentially independent rules against the entire training set, the protection of filtering against overtraining remains stronger. Giving up the power to synthesize new rules thus provides an overtraining payoff as well as a leverage one.

### Rule Interdependence

While the connections between rules in a rule sequence are more limited than the inheritance of rule ancestors found in decision trees, it is still interesting to be able to characterize and quantify the rule dependencies that are present. We have therefore added code that keeps track, whenever a rule is applied at a site, of a dependency tree showing the earlier rule applications that that rule depends on. For example, the dependency tree from the Brown Corpus data in Fig. 6 shows a case where the last rule that applied at this particular site (the bottom line in the figure, representing the root of the tree), which changed JJ to RB, depended on earlier rules that changed the previous site (relative position −1) to VBN and the following one (position +1) to DT. (The final number on each line tells during what pass that rule was learned. While recorded internally as trees, these structures actually represent dependency DAGs, since one rule application may be an ancestor of another along more than one path.) All sites start out assigned a null dependency tree representing the baseline heuristic choice. The application of a rule causes a new tree to be built, with a new root node, whose children are the dependency trees for those neighboring locations referenced by the rule pattern. At the end of the training run, the final dependency trees are sorted, structurally similar trees are grouped together, and the classes are then sorted by frequency and output along with the list of rules learned.

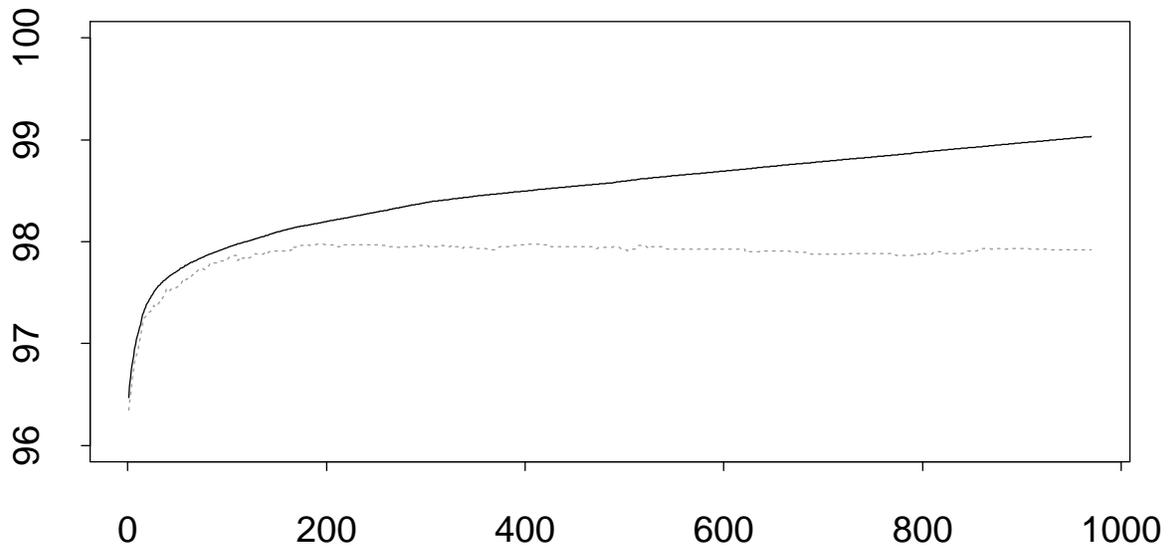

Figure 3: Training Set (solid line) and Test Set (dotted line) Performance on Greek Corpus

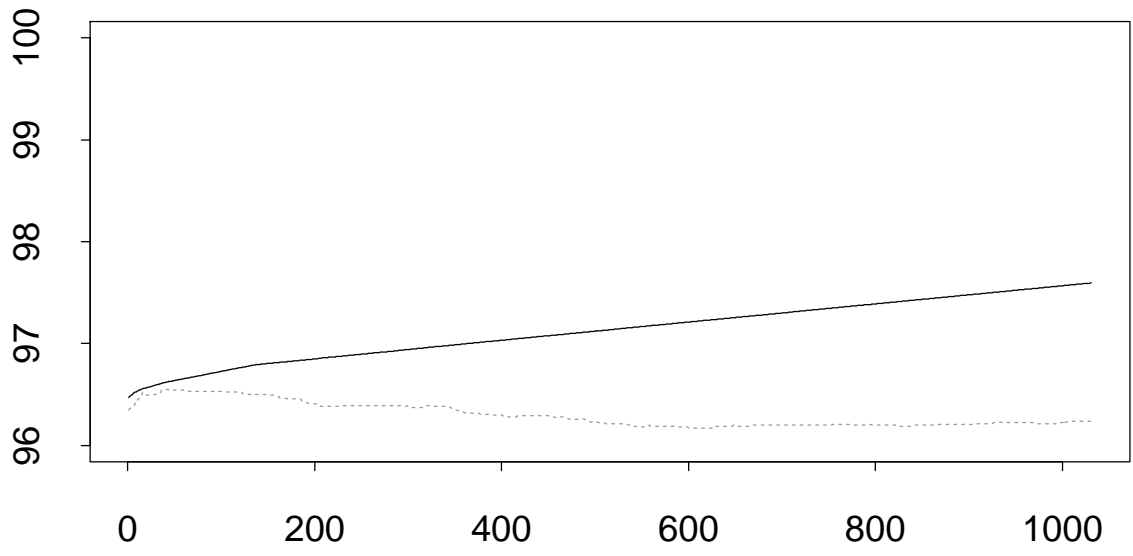

Figure 4: Training with 1 Irrelevant Template on Greek Corpus

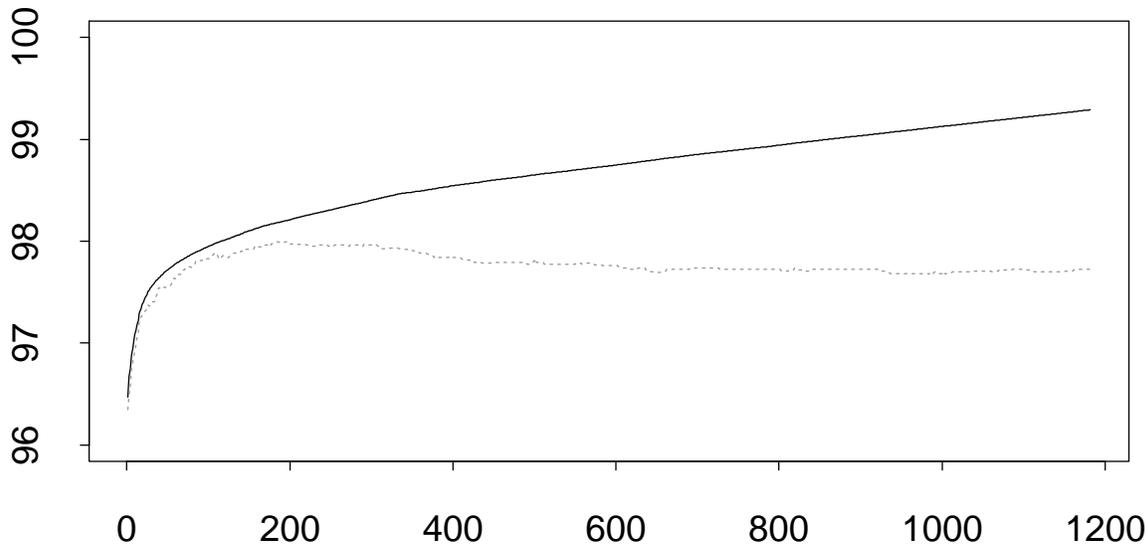

Figure 5: Training with 7 Relevant and 1 Irrelevant Templates

Certain common classes of dependency can be noted in the resulting trees. Correction rules result when one rule makes an overly general change, which affects not only appropriate sites, but also inappropriate ones, so that a later rule in the sequence undoes part of the earlier effect. One dependency of this type from our Brown Corpus run can be seen in Fig. 7. Here the first rule was the more general one that changed PP$ to PPO whenever it follows VBD. While that rule was generally useful, it overshot in some cases, causing the later learning of a correction rule that changed PPO back to PP$ after RB VBD.

Chaining rules occur in cases where a change ripples across a context, as in Fig. 8. The first rule to apply here (21) changed QL to AP in relative position +2. That change enabled the RB to QL rule (181) at position +1, and together those two changes enabled the root rule (781). Note that this two-step rule chain has allowed this rule to depend indirectly on a current tag value that is further away than could be sensed in a single rule, given the current maximum template width.

The dependency tree output also shows something of the overall degree and nature of rule interdependence. The trees for a run on 50K words of the Brown Corpus bear out that rule dependencies, at least in the part-of-speech tagging application, are limited. Of a total of 3395 sites changed during training, only 396 had dependency trees with more than one node, with the most frequent such tree appearing only 4 times. Thus the great majority of the learning in this case came from templates that applied in one step directly to the baseline tags, with leveraging being involved in only about 12% of the changes.

The relatively small amount of interaction found between the rules also suggests that the order in which

```
+1:     —     —         CD/DT  NN   —    (7)
−1:  —  HVD  VBD/VBN    —            —    (8)
 0:  —   —   VBN        JJ/RB  DT   —    (649)
```

Figure 6: Sample Dependency Tree from Brown Corpus Data

```
0:  —   VBD  PP$/PPO  —  —  (30)
0:  RB  VBD  PPO/PP$  —  —  (174)
```

Figure 7: Sample Correction Class Dependency Tree from Brown Corpus Data

```
+2:  —   —    —        QL/AP  CS   —   (21)
+1:  —   —    RB/QL    AP     CS       (181)
 0:  —   —    NNS/VBZ  QL     AP       (781)
```

Figure 8: Sample Chaining Class Dependency Tree from Brown Corpus Data

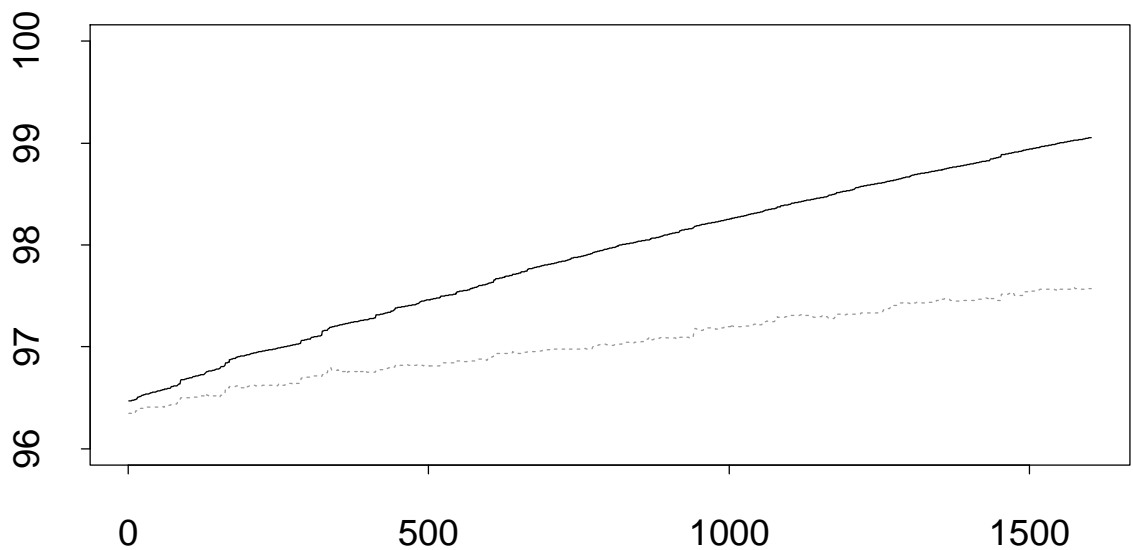

Figure 9: Training and Test Set Performance on Greek, Random Rule Choice

the rules are applied is not likely to be a major factor in the success of the method for this particular application, and initial experiments tend to bear this out. Fig. 3 earlier showed a training run on Greek text using the largest net benefit choice rule that Brill proposes. Note that, on this Greek corpus, the initial baseline level of choosing the most frequent training set tag for each word is already quite good; performance on both sets further improves during training, with most of the improvement occurring in the first few passes. In comparison, Fig. 9 gives the results for a training run where the next rule at each step was randomly selected from among all rules that had a net positive effect of any size. While the progress is more gradual, both the training and test curves reach very close to the same maxima under these conditions as they do when the largest net benefit rule is chosen at each step. Note that it does take more rules to reach those levels, since the random training frequently chooses more specific rules that would have been subsumed by more general ones chosen later. Thus the largest net benefit ranking criterion is a useful one, particularly if one wants to find a short initial subsequence of rules which achieves the bulk of the good effect. But at least for this task, where there is little interdependence, choice of search order does not much affect the final performance achieved.

## Future Work

The general analysis of rule sequences in relation to decision trees presented here is based on experiments primarily in the part-of-speech tagging domain. Within that domain, it would be useful to quantify more clearly whether or not rule sequence learning is more effective than traditional decision tree methods when applied to the same corpora and making use of the same factors. Such experiments would better illuminate the trade-offs between the ability to combine predicates into more complex rules on the one hand and the ability to leverage partial results and resist overtraining on the other. It would also be useful to test the data presented here on overtraining risk and on rule interdependence in other domains, particularly ones where the degree of rule interdependence could be expected to be greater. Further exploration of the connections between rule sequences and decision trees may also suggest other approaches, perhaps blends of the two, that would work better in some circumstances.

Within rule sequence learning itself, other ranking schemes for selecting the next rule to apply might be able to improve on the simple maximum net benefit heuristic. We are currently exploring the use of likelihood ratios for this purpose. It may also be possible to control for the remaining risk of overtraining in a more sensitive way than with a simple threshold. Decision trees often use selective pruning to control overtraining, and deleted estimation (Jelinek and Mercer, 1980) or other cross-validation techniques are also natural suggestions for this purpose, but it is difficult to see how to apply any of these techniques to bare rule sequences because they contain hidden dependencies between rules, so that there is no obvious way to delete selected rules or to interpolate between two different rule sequences. One goal for collecting the dependency tree data is to make it possible to prune or restructure rule sequences, using the recorded dependencies to maintain consistency among the remaining rules.

## Conclusions

Transformational rule sequence learning is a simple and powerful mechanism for capturing the patterns in linguistic data, which makes it an attractive alternative well worth further exploration. Brill has showed that its performance for part-of-speech tagging can surpass that of the HMM models most frequently used, while producing a more compact and perhaps more interpretable model.

While its results can be compared with those of HMM models, the rule sequence technique itself seems to have more in common with decision trees, especially in its ability to automatically select at each stage from a large space of possible factors the predicate or rule that appears to be most useful. Decision trees synthesize complex rules from elementary predicates by inheritance; rule sequence learning, on the other hand, prespecifies in the templates essentially the full space of possible rules, with each rule acting largely independently. This restriction in power turns out not to be crippling as long the template set can be made rich enough to cover the patterns likely to be found in the data, and it brings two important benefits in return: first, breaking the learning process into independent rules means that they can be applied to the whole corpus as they are learned, so that where neighboring patterns in the data are interrelated, the rules can leverage off the best estimates regarding their surroundings; and second, since the independent rules continue to be scored against the whole training corpus, a substantial measure of protection against overtraining compared to decision trees is gained.

## References

Breiman, Leo, Jerome H. Friedman, Richard A. Olshen, and Charles J. Stone. 1984. *Classification and Regression Trees*. Pacific Grove, California: Wadsworth & Brooks/Cole.

Brill, Eric. 1992. A simple rule-based part of speech tagger. In *Proceedings of the DARPA Speech and Natural Language Workshop, 1992*.

Brill, Eric. 1993a. Automatic grammar induction and parsing free text: A transformation-based approach. In *Proceedings of the DARPA Speech and Natural Language Workshop, 1993*.

Brill, Eric. 1993b. *A Corpus-Based Approach to Language Learning*. Ph.D. thesis, University of Pennsylvania.

Brill, Eric. 1994. A report of recent progress in transformation-based error-driven learning. In *Proceedings of the ARPA Workshop on Human Language Technology, March, 1994*.

CATSS. 1991. Produced by Computer-Assisted Tools for Septuagint Studies, available through the University of Pennsylvania's Center for Computer Analysis of Texts.

Church, Kenneth. 1988. A stochastic parts program and noun phrase parser for unrestricted text. In *Second Conference on Applied Natural Language Processing*. ACL.

Cutting, D., J. Kupiec, J. Pederson, and P. Sibun. 1992. A practical part-of-speech tagger. In *Proceedings of the Third Conference on Applied Natural Language Processing*. ACL.

DeRose, Steven J. 1988. Grammatical category disambiguation by statistical optimization. *Computational Linguistics*, 14(1):31–39.

Francis, W. Nelson and Henry Kucera. 1979. Manual of information to accompany a standard corpus of present-day edited American English, for use with digital computers. Technical report, Department of Linguistics, Brown University.

Jelinek, F. 1985. Markov source modeling of text generation. In ed. J.K. Skwirzinski, editor, *Impact of Processing Techniques of Communication*. Nijhoff, Dordrecht.

Jelinek, F. and R. L. Mercer. 1980. Interpolated estimation of Markov source parameters from sparse data. In E.S. Gelsema and L. N. Kanal, editors, *Pattern Recognition in Practice*. North-Holland, Amsterdam, pages 381–397.

Quinlan, J. Ross. 1993. *C4.5: Programs for Machine Learning*. Morgan Kaufmann.

Quinlan, J. Ross and Ronald L. Rivest. 1989. Inferring decision trees using the minimum description length principle. *Information and Computation*, 80:227–248.

Weischedel, Ralph, Marie Meteer, Richard Schwartz, Lance Ramshaw, and Jeff Palmucci. 1993. Coping with ambiguity and unknown words through probabilistic methods. *Computational Linguistics*, 19(2):359–382.